\documentclass{article}

\usepackage{microtype}
\usepackage{graphicx}
\usepackage{subcaption}
\usepackage{booktabs} 
\usepackage{xcolor}
\usepackage{multirow}
\usepackage{adjustbox}

\usepackage{hyperref}

\newif\ifshow
\showtrue
\showfalse

\ifshow
\newcommand{\Cody}[1]{\textcolor{blue}{[\textbf{Cody:} #1]}}
\newcommand{\Matthew}[1]{\textcolor{orange}{[\textbf{Matthew:} #1]}}
\newcommand{\todo}[1]{\textcolor{red}{[TO DO: #1]}}
\else
\newcommand{\Cody}[1]{}
\newcommand{\Matthew}[1]{}
\newcommand{\todo}[1]{}
\fi

\usepackage[accepted]{mlsys2023}

\mlsystitlerunning{Reduce, Reuse, Recycle: Improving Training Efficiency with Distillation}

\begin{document}

\twocolumn[
\mlsystitle{Reduce, Reuse, Recycle: Improving Training Efficiency with Distillation}



\mlsyssetsymbol{equal}{*}

\begin{mlsysauthorlist}
\mlsysauthor{Cody Blakeney}{txst,mosaic}
\mlsysauthor{Jessica Zosa Forde}{brown,mosaic}
\mlsysauthor{Jonathan Frankle}{mosaic}
\mlsysauthor{Ziliang Zong}{txst}
\mlsysauthor{Matthew L. Leavitt}{mosaic}
\end{mlsysauthorlist}

\mlsysaffiliation{txst}{Department of Computer Science, Texas State University, San Marcos, Texas, USA}
\mlsysaffiliation{brown}{Brown University, Providence, Rhode Island, USA}
\mlsysaffiliation{mosaic}{MosaicML, San Francisco, California, USA}

\mlsyscorrespondingauthor{Matthew L. Leavitt}{matthew@mosaicml.com}
\mlsyscorrespondingauthor{Jonathan Frankle}{jonathan@mosaicml.com}

\mlsyskeywords{Machine Learning, MLSys}

\vskip 0.3in

\begin{abstract}

Methods for improving the efficiency of deep network training (i.e. the resources required to achieve a given level of model quality) are of immediate benefit to deep learning practitioners. Distillation is typically used to compress models or improve model quality, but it's unclear if distillation actually improves training \emph{efficiency}. Can the quality improvements of distillation be converted into training speed-ups, or do they simply increase final model quality with no resource savings? We conducted a series of experiments to investigate whether and how distillation can be used to accelerate training using ResNet-50 trained on ImageNet and BERT trained on C4 with a masked language modeling objective and evaluated on GLUE, using common enterprise hardware (8x NVIDIA A100). We found that distillation can speed up training by up to 1.96x in ResNet-50 trained on ImageNet and up to 1.42x on BERT when evaluated on GLUE. Furthermore, distillation for BERT yields optimal results when it is only performed for the first 20-50\% of training. We also observed that training with distillation is almost always more efficient than training without distillation, even when using the poorest-quality model as a teacher, in both ResNet-50 and BERT. Finally, we found that it's possible to gain the benefit of distilling from an ensemble of teacher models, which has $O(n)$ runtime cost, by randomly sampling a single teacher from the pool of teacher models on each step, which only has a $O(1)$ runtime cost. Taken together, these results show that distillation can substantially improve training efficiency in both image classification and language modeling, and that a few simple optimizations to distillation protocols can further enhance these efficiency improvements.

\end{abstract}
]



\printAffiliationsAndNotice{}  

\section{Introduction}
Neural network training has a waste problem. While considerable attention has been given to the cost, electricity consumption, and carbon footprint of training large neural networks, reported figures often dramatically under-represent the true cost of training because they focus only on the final training run. However, far more resources are usually dedicated to \emph{searching} for the optimal hyper-parameters for a given model, dataset, or recipe of techniques. Models trained during this search phase are typically used to reduce the hyperparameter search space, and as such serve as signposts for what \emph{not} to do. What these previously-trained models have learned is not usually leveraged to directly improve training. This leads us to define the \emph{Iterated Runs Problem}: How can previous training runs be used to improve the efficiency\footnote{We define efficiency here as in \citet{blalock_evaluating_2021}: achieving a target level of model quality using fewer resources (e.g. GPU hours) than a baseline, or achieving an increased level of model quality using the same resources as a baseline.} of a subsequent training run?

A number of machine learning research areas—some of them seemingly quite disparate—address the Iterated Runs Problem, including sample pruning and core-set selection \cite{vodrahalli_are_2018,toneva_forgetting_2019,swayamdipta_cartography_2020,coleman_coreset_selection_2020,chitta_training_subset_2020,feldman_what_memorize_2020,paul_deep_2021,mindermann_rhols_2022,sorscher_beyond_scaling_2022}, active learning \cite{ren_survey_active_2021}, model averaging \cite{wortsman_robust_averaging_2021,matena_merging_2022,wortsman_soup_2022}, and knowledge distillation (KD) \cite{hinton2015distilling,gou_knowledge_2021}. Work in these areas is not always conducted or framed with goal of improving training efficiency. Some approaches are scientifically valuable but impractical—for example, using \citet{feldman_what_memorize_2020}'s memorization and influence scores for data pruning requires training a number of models equal to the size of training dataset. Similarly, knowledge distillation is often studied with the explicit goal of either compressing models or maximizing model quality (without regard for resource usage), and is rarely studied as a way to improve training efficiency.

The work that \emph{has} chosen to study distillation from the perspective of training efficiency typically quantifies "efficiency" in units of optimization steps\cite{yim2017gift,yang_snapshot_2018,furlanello2018born,lin_ensemble_2020,liu_knowledge_efficient_2022}. This completely overlooks the increased computational burden of distillation and real-time cost of training (accelerator resources are typically priced in units of time, not optimization steps). Thus it is presently unclear whether distillation can be leveraged to improve resource usage efficiency.

Framing knowledge distillation as a solution to the Iterated Runs Problem and examining it from the perspective of improving training efficiency leads to a number of interesting questions. First and foremost, can distillation be used to improve training efficiency, or do its computational costs outweigh its benefits? And can we optimize traditional distillation paradigms to reduce resource usage while retaining model quality improvements? For example, traditional distillation paradigms distill for all of training, but is this truly necessary to obtain the benefits of distillation? Previous work has also shown that distilling from ensembles of models can yield benefits beyond those of distilling from a single model \cite{furlanello2018born,zhang_be_2019, wang_efficient_checkpoints_2022, liu_knowledge_efficient_2022}, but distilling from multiple models is especially computationally costly. Can we obtain the benefits of having multiple teacher models without having to pay the full computational cost? And when sweeping across hyperparameters one often ends up with suboptimal models. Do suboptimal models need to be discarded, or can they make useful teachers? And does using distillation to accelerate hyperparameter search change the optimal hyperparameter choice?

We conducted a series of experiments to investigate the utility of distillation for improving training efficiency in an Iterated Runs scenario using ResNet-50 \cite{he2016deep} trained on ImageNet \cite{russakovsky2015imagenet} and BERT \cite{devlin2018bert} trained on C4 \cite{2019t5} with a masked language modeling objective and evaluated on GLUE \cite{wang_glue_2019}. In these experiments, we conducted a hyperparameter sweep across four learning rate values, then used one (or more) of the trained models as a teacher(s) to distill a fifth model of the same architecture, and report the following results\footnote{All experiments were conducted on 8x NVIDIA A100 accelerators}:

\vspace{-3mm}
\newcommand{\contspace}{\vspace{-1mm}}
\begin{itemize}
    \item Distillation improves training efficiency. We found that distillation can speed up training by up to 1.96x in ResNet-50 trained on ImageNet, up to 1.20x on BERT when evaluated on masked pretraining accuracy, and up to 1.42x on BERT when evaluated on GLUE.
    \contspace
    \item Distillation schedules matter. Distillation for BERT yields optimal results when it is only performed for the first 20-50\% of training; training BERT with distillation for the entirety of training actually \emph{decreases} efficiency. In contrast, distilling for the entirety of training is optimal for ResNet-50 on ImageNet.
    \contspace
    \item Model quality does not consistently predict teacher quality. Training with distillation is almost always more efficient than training without distillation, even when using the poorest-quality model as a teacher, in both ResNet-50 and BERT.
    \contspace
    \item Randomly sampling one teacher model from a pool of teachers on each iteration provides similar quality gains as those obtained from using that same pool of models as an ensemble of teachers on every iteration in ResNet-50 trained on ImageNet. This effectively reduces the runtime cost of teacher ensembles from $O(N)$ to $O(1)$.
    \contspace
    \item We observed differences between mean squared error (MSE) and KL-Divergence (KL) distillation loss in ResNet-50 trained on ImageNet. MSE is more robust—it more consistently yields higher quality student models across a wide range hyperparameter values—but KL-Divergence distillation loss yields the \emph{best} student models.
    \contspace
\end{itemize}
\vspace{-2mm}

These results show that distillation can substantially improve training efficiency in both image classification and language modeling. Furthermore, our results show that distillation is a consistently safe bet for improving the efficiency of training, regardless of quality of the teacher model. We also show that the benefits of distillation on training speed and model quality are fungible. This means that our proposed optimizations to distillation protocols—randomly sampling from ensembles of teacher models, and distilling for the beginning ~30\% of training (in BERT)—can be flexibly used to improve model quality or reduce training costs, depending on the needs of the practitioner. Taken together, this work emphasizes the utility of distillation for improving the efficiency of training deep neural networks.

\section{Related Work}

\subsection{Knowledge Distillation}

Knowledge distillation \cite{hinton2015distilling} is a well-established practice for model compression and improving the quality of models \cite{gou_knowledge_2021}. Distillation is often an ingredient in training recipes that push the limits of model quality \cite{xie2020self,touvron_deit_2021,beyer_patient_distillation_2021}. Teacher models are traditionally larger than student models, though too large of a size discrepancy between teacher and student models can reduce the efficacy of distillation \cite{mirzadeh2020improved_assistant}. As such, numerous approaches have been proposed to more effectively utilize models of similar or identical architecture as teachers \cite{yim2017gift,yang_snapshot_2018,furlanello2018born,zhang_be_2019,wang_efficient_checkpoints_2022}.

\subsection{Teacher Ensembling and Self-Distillation}

Of particular relevance to the Iterated Runs Problem are approaches that use self-distillation—distilling from previous checkpoints in a training run \cite{yang_snapshot_2018,furlanello2018born,zhang_be_2019,xu_bert_self_distill_ensemble_2020}—and distilling from ensembles of teachers, which has been shown to be particularly effective for improving model quality \cite{malinin_ensemble_2019,zhang_be_2019,lin_ensemble_2020,asif_ensemble_2020,xu_bert_self_distill_ensemble_2020,allen-zhu_towards_understanding_ensemble_2021,wang_efficient_checkpoints_2022,liu_knowledge_efficient_2022}. The work of \citet{gontijo-lopes_no_one_representation_ensembling_2021} shows that the benefits of ensembles seem to be due to their response diversity, though they did not examine distillation specifically. Some approaches combine both self-distillation and ensembling \cite{zhang_be_2019,xu_bert_self_distill_ensemble_2020,wang_efficient_checkpoints_2022}.

\subsection{Distillation in Language Models}

While many studies have examined the utility of distillation for improving model quality in vision models, comparatively few studies have studied this phenomenon in language models \cite{xu_bert_self_distill_ensemble_2020}. The majority of research into distillation for language models appears to focus on the problem of compression, not quality \cite{sun_patient_2019, jiao_tinybert_2020,liu_fastbert_2020,sanh_distilbert_2020,xu_bert--theseus_2020,zhang_ternarybert_2020}.

\subsection{Distillation for Stepwise Training Speedups}

A number of works have claimed that distillation improves training efficiency based on results demonstrating that distillation can reduce the number of optimization steps necessary to achieve a given level of model quality compared to a baseline model \cite{yim2017gift,yang_snapshot_2018,furlanello2018born,lin_ensemble_2020,liu_knowledge_efficient_2022}. Of particular interest is \citet{liu_knowledge_efficient_2022}, who showed that much of the benefit of distillation can be obtained at reduced computational cost by distilling intermittently (i.e. once every K steps). They also showed that randomly sampling one teacher model from a pool of possible teachers on each step is nearly as effective as distilling from the entire ensemble of teachers. Unfortunately, \citet{liu_knowledge_efficient_2022}, along with most previous work claiming that distillation improves training efficiency, quantify "efficiency" in units of optimization steps. This completely overlooks the increased computational cost of distillation, leaving it unclear whether any of these approaches truly improve resource usage efficiency.

\section{Methodology}

We designed a series of experiments examining the utility of distillation for improving training efficiency. We define efficiency here as in \citet{blalock_evaluating_2021}: achieving a target level of model quality using fewer resources (e.g. GPU hours) than a baseline, or achieving an increased level of model quality using the same resources as a baseline. Our experiments address the following key questions:

\begin{itemize}
    \item Can distillation improve training efficiency?
    \item Should you distill for all of training?
    \item Do sub-optimal models make bad teachers
    \item Are more teachers helpful?
\end{itemize}  

\subsection*{How do we distill?}

For all our experiments we use models with exactly the same architectures, optimizers, and datasets. We are \emph{not} attempting to compress the knowledge of a larger model into a smaller one but instead attempting to train the \emph{same} model to a the same or higher quality with fewer resources.

While there are many methods for performing knowledge distillation. In this work we chose to using only response based distillation making no use of internal model features. This is the most flexible and allows us to compare approaches across domain and model arichtecture. We compare the use of both Kullback-Leibler (KL) divergence loss $\mathcal{L}_{kl}$ as described in\cite{hinton2015distilling} and MSE loss $\mathcal{L}_{MSE}$. Student models are trained using a linear combination of either of the KD losses $\mathcal{L}_{kd}$ as cross entropy loss $\mathcal{L}_{ce}$. 

$$\mathcal{L} = \lambda \mathcal{L}_{kd} + \mathcal{L}_{ce}$$

Where $\lambda$ controls the weight of the KD loss term. We note that we use the term ``distillation" even though the student and teacher model architectures are identical within an experiment. We are using distillation to examine whether the teaching signal can be used to improve training efficiency, \emph{not} to compress models.

\subsection*{How do we evaluate?}

Our primary concern is wall-clock efficiency. For Image classification we measure that efficiency as speedup to reach the same Top-1 accuracy as the baseline. For BERT we evaluate both Masked Language Modeling (MLM) accuracy when distilling as well as performance of the pretrained model on downstream tasks from the GLUE Benchmark \cite{wang_glue_2019}.

\section{Experimental Setup}

\subsection{How do we Train?}

\label{lab:train}

\subsubsection{Data and Models}
We perform model distillation on two domains and tasks. Image classification using the ResNet-50 \cite{he2016deep} architecture on the ImageNet \cite{russakovsky2015imagenet} dataset and Masked Language Modeling Pre-training with BERT \cite{devlin2018bert} on the C4 \cite{2019t5} dataset. 


For ResNet-50 training on ImageNet we follow the basic precedure described by \cite{he2016deep} using standard 224 x 224 test resolution. The only notable exception being the use of SGDW \cite{loshchilov2017decoupled} for the optimizer and cosine annealing \cite{LoshchilovH_sgdr_17} as the learning rate scheduler. 

For BERT pre-training on C4 we use AdamW \cite{loshchilov2017decoupled} and linear decay. Instead of a fixed warmup length we scale the warmup period by the percentage of training duration. More details can be seen in table \ref{tab:shared_hparams}. 

\subsubsection{Hardware and trainer}
We conducted our experiments using Composer \cite{tang_composer_2022} a PyTorch \cite{paszke_pytorch_2019} library for efficient training. Our plots are visualized using Seaborn \cite{waskom_seaborn_2021}. All experiments were conducted on 8x NVIDIA A100 80gb.

\begin{table}[h]
\centering
\caption{Shared Training hyperparameters for teacher models and students.}
\label{tab:shared_hparams}
\adjustbox{max width=\columnwidth}{%
\begin{tabular}{@{}lrr@{}}
\toprule
Model               & ResNet-50            & BERT                     \\ \midrule
Batch size          & 2048                 & 4096                     \\
Training Duration (baseline model) & 90 epochs & 286.72M sequences   \\
Max Sequence Len    & N/A                  & 128                      \\
Optimizer           & SGDW                 & AdamW                    \\
Weight Decay        & 5.00E-04             & 1.00E-05                 \\
Momentum            & 0.875                & N/A                      \\
Warmup              & 8 epochs             & 6\% of training duration \\
Scheduler           & Cosine Annealing     & Linear Decay             \\ \bottomrule
\end{tabular}}

\caption{Training hyperparameters used for teacher models.}
\adjustbox{max width=\columnwidth}{%
\begin{tabular}{@{}ccccc@{}}
\toprule
\multicolumn{1}{l}{Hyper Parameters} & \multicolumn{4}{c}{Model}                 \\ \midrule
ResNet                                                       & B1       & B2       & B3       & B4       \\
lr                                                           & 1        & 2.045    & 0.01     & 0.1      \\ \midrule
BERT                                                         & A1       & A2       & A3       & A4       \\
lr                                                           & 5.00E-04 & 5.00E-04 & 1.00E-04 & 1.00E-04 \\
wd                                                           & 1.00E-05 & 1.00E-04 & 1.00E-05 & 1.00E-04 \\ \bottomrule
\end{tabular}}
\end{table}

\begin{table*}
\caption{Results of hyper parameter sweep of teacher models on both ResNet-50 ImageNet and BERT on C4 and KD. The teacher model is the highest-quality model (B1 for ResNet-50, A1 for BERT) at standard training length (90 epochs for ResNet-50, 286.72M sequences for BERT). KL: KL-Divergence distillation loss; MSE: Mean-squared error distillation loss.}
\label{tab:teacher-val-naive-kd}
\centering
\adjustbox{max width=\linewidth}{%
\begin{tabular}{llrrrrrr}
\toprule
               & Model &            B1 &            B2 &            B3 &            B4 &       KD - KL &      KD - MSE \\
{ResNet-50 - ImageNet} & Epochs &               &               &               &               &               &               \\
\midrule
\multirow{5}{*}{Top-1 Val Accuracy} & 22  &      72.31\% &      72.96\% &      39.53\% &      68.16\% &      74.82\% &      75.64\% \\
               & 45  &      75.43\% &      75.64\% &      60.85\% &      73.46\% &      76.51\% &      76.73\% \\
               & 90  &      76.62\% &           76.40\% &      68.85\% &      75.86\% &      77.333\% &      77.21\% \\
               & 135 &      76.79\% &      76.52\% &      71.27\% &      76.85\% &      77.42\% &      77.31\% \\
               & 180 &      76.87\% &      76.51\% &      72.29\% &      77.17\% &      77.53\% &      77.37\% \\
\toprule
             & Model &  A1 &  A2 &  A3 &  A4 & KD - KL & \\
{BERT - C4 MLM} & Training Steps &           &              &              &              \\
\midrule
\multirow{5}{*}{MLM Val Accuracy} & 17500 &  63.11\% &     62.59\% &     53.38\% &     48.89\% & 64.94\% & \\  
             & 35000 &  65.49\% &     64.77\% &     58.52\% &     51.54\% & 66.71\% & \\ 
             & 52500 &  66.59\% &     65.79\% &     61.13\% &     52.28\% & 67.43\% & \\ 
             & 70000 &  67.31\% &     66.41\% &     62.71\% &     52.40\% & 67.88\% & \\ 
             & 87500 &  67.77\% &     66.77\% &     63.79\% &     52.49\% & 68.23\% & \\
\bottomrule
\end{tabular}}


\end{table*}

\begin{figure*}[ht]
     \centering
     \vfill
          \begin{subfigure}[t]{0.49\textwidth}
         \centering
         \includegraphics[width=\textwidth]{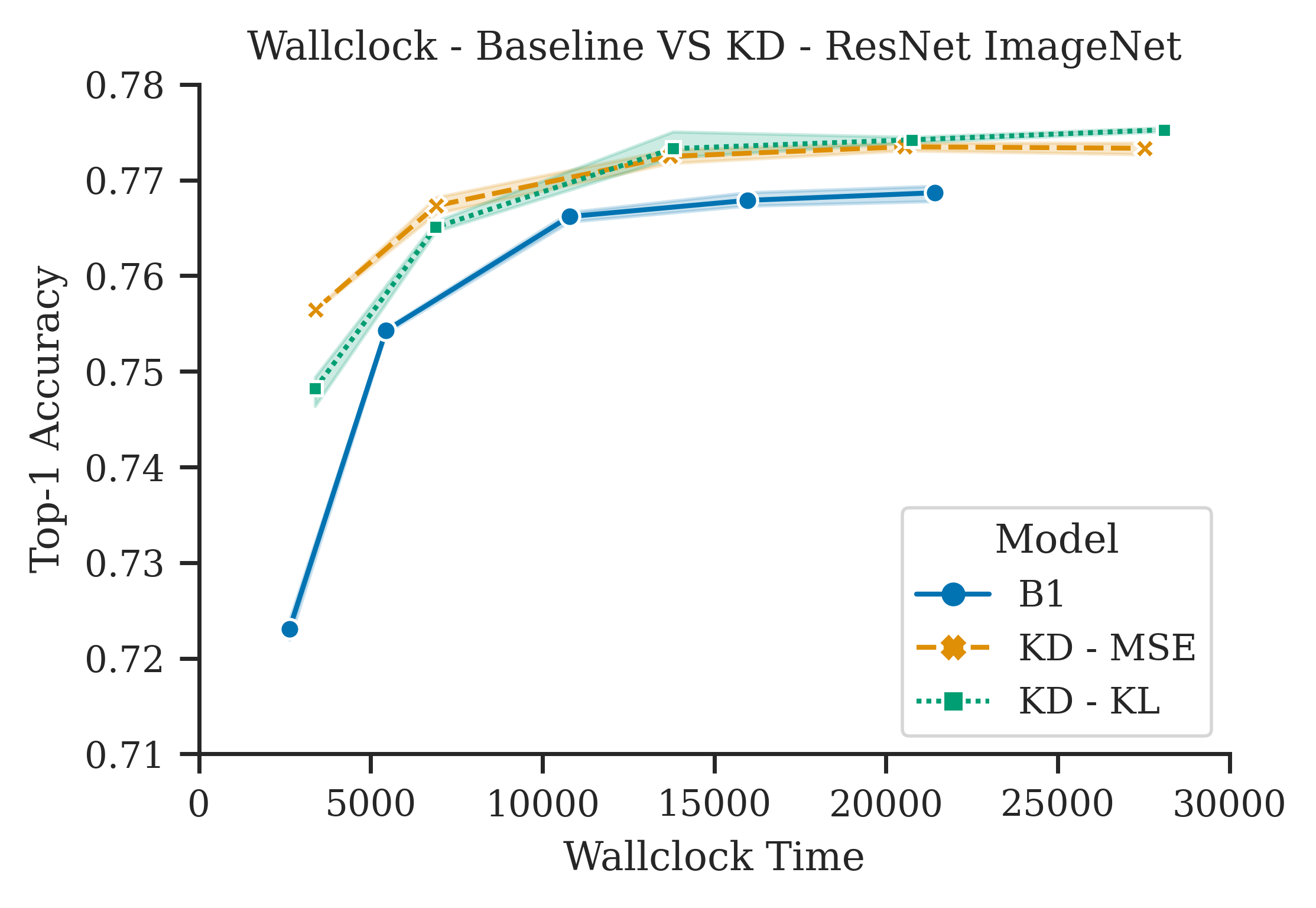}
         \caption{ResNet-50 trained on ImageNet with vs. without distillation. Wallclock time-to-train (x-axis) comparison of Teacher model B1 vs student models trained with B1 as teacher using MSE and KL Div losses. Individual points along each line denote models trained for the number of epochs reported in Table 3.}
         \label{fig:simple-kd-wallclock}
     \end{subfigure}
     \hfill
     \begin{subfigure}[t]{0.49\textwidth}
         \centering
         \includegraphics[width=\textwidth]{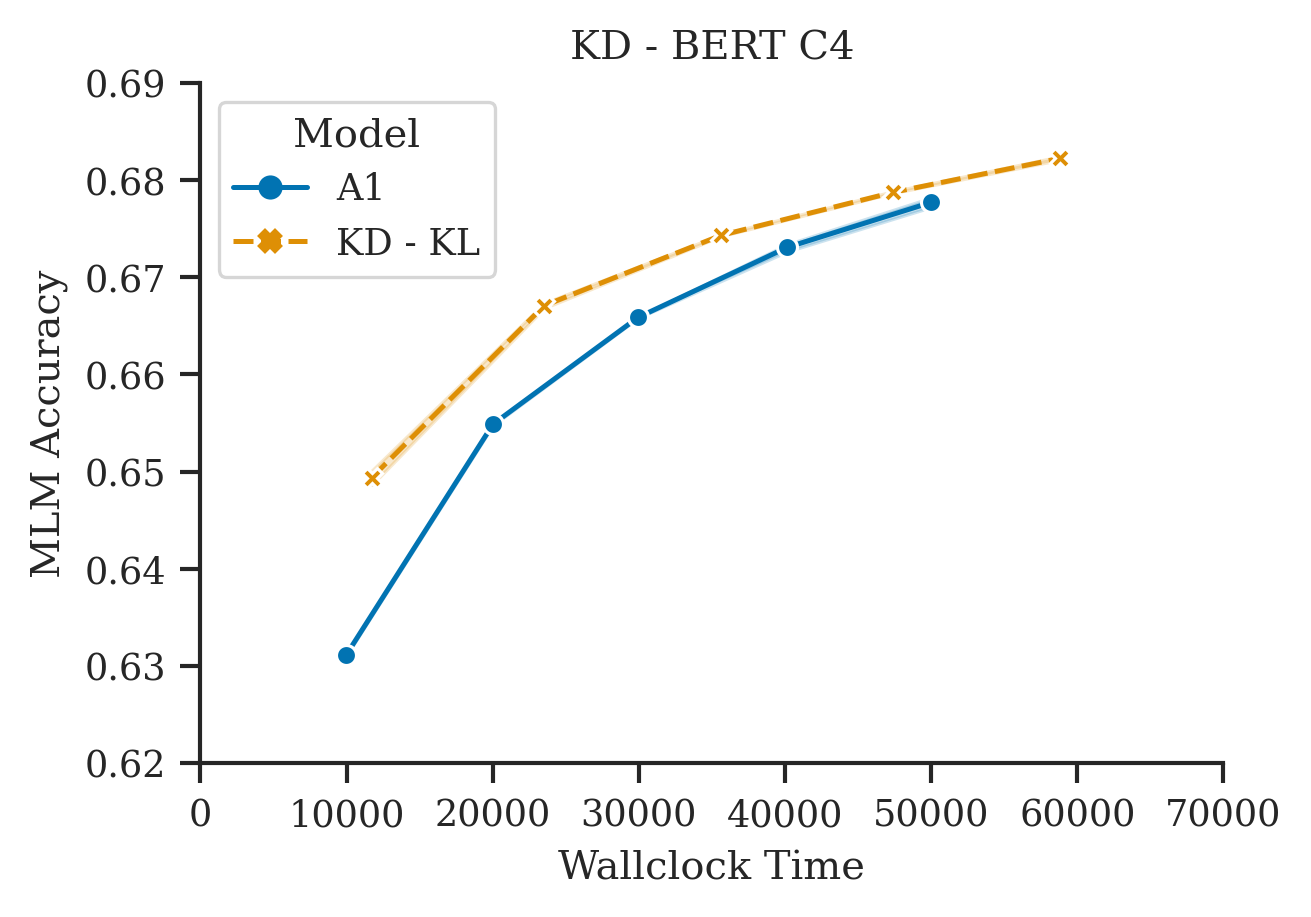}
         \caption{MLM Accuracy of pretraining on C4 dataset with BERT utilizing Early-phase-only distillation. Individual points along each line denote models trained for the number of sequences reported in Table 3.}
         \label{fig:bert-pareto}
     \end{subfigure}

        \caption{Wallclock comparisons applying knowledge distillation to ResNet and BERT.}
        \label{fig:simple-distillation}
\end{figure*}

\section{Results}

\subsection{Simple distillation can improve training efficiency}
\label{sec:simple}
While distillation can improve the final quality of a trained model, it's unclear whether distillation can actually improve training \emph{efficiency}. We define efficiency as in \citet{blalock_evaluating_2021}: achieving a target level of model quality using fewer resources than a baseline, or achieving an increased level of model quality using the same resources as a baseline. In all of the experiments presented in this work, the resource is wall-clock time-to-train on 8x NVIDIA A100 (see Methodology: Hardware), and our baselines are ResNet-50 \cite{he2016deep} trained on ImageNet \cite{russakovsky2015imagenet} and BERT \cite{devlin2018bert} trained on C4 (\citet{2019t5}; see Section \ref{lab:train}).

We wanted our experiments to be simple, but also reflective of typical machine learning workflows as they pertain to the Iterated Runs Problem (see Introduction). This led us to design the following experiment to determine whether distillation can improve training efficiency: We first conducted a hyperparameter sweep by sweeping across five learning rate values (all other hyperparameters were kept constant; see Methodology: Models and Datasets). We then selected the highest-quality model from the sweep as baseline, and also as the teacher model to train a student model with distillation. We then compared the wall clock time-to-train for the baseline model to reach its final eval accuracy to the wall clock time-to-train for the distilled model to reach the baseline model's final eval accuracy.

If distillation does not improve training efficiency, we expect the distilled model to take longer than the baseline model to reach the baseline model's final accuracy. Alternatively, if distillation improves training efficiency, then we expect the distilled model to reach the baseline final accuracy faster than the baseline model.

In ResNet-50 trained on ImageNet, we found that distillation substantially improves the training outcome, but not efficiency. The baseline model (B1) reached 76.6\% eval accuracy in 179.8 minutes at the end of training for 90 epochs, while the distilled model achieved the same accuracy in 200  minutes at epoch 80 of training out of 90, a wallclock slow down of around 11\%. It however finishes a final eval accuracy of 77.2\% 


While this experiment demonstrates that distillation may not result in a wall-clock speedup, it fails to leverage the fact that model quality and training speed can be fungible. If speed and quality are fungible, then a quality improvement can be converted into a speed-up by training for less time. We can thus trade the excess quality of our distilled model relative to our baseline for additional speed-up. Accordingly, we trained the distilled model for 45 epochs (and scaled the learning rate decay accordingly) and found that we were able to reach the final accuracy of the baseline model (which was trained for 90 epochs) in 91.6 minutes, a 1.96x speed-up (Figure \ref{fig:simple-kd-wallclock}).


Our previous experiments showed that simple distillation can speed up training ResNet-50 on ImageNet by up to 1.96x, but it's possible that distillation's efficiency improvements are specific to the specific experiment configuration. In order to determine whether distillation is more broadly practical for improving training efficiency, we repeated the same set of experiments in BERT trained on C4 (see Methodology: Models and Datasets). We trained our baseline model on 286,720,000 sequences of 128 tokens, which took 11.3 hours, and achieved 67.31\% val MLM accuracy and 83.37\% accuracy on GLUE. The distilled model achieved baseline MLM accuracy in 18 hours. 


The overhead per step for distillation with BERT is significantly higher (~60\% vs ~30\%). Even when accounting for shorter training regimes we find that although distillation with BERT can be a stepwise improvement, because of the additional overhead there is no case where it is an efficiency improvement.

Our experiments demonstrate that distillation can substantially improve training efficiency for ResNet-50 trained on ImageNet, but that distillation may not improve training efficiency for BERT trained on C4.
However, it's possible that our distillation configuration is sub-optimal, and there are potential efficiency gains that remain to be realized.

We also repeated these experiments using MSE loss student-teacher loss instead of KL-divergence/student-teacher loss. For ResNet on ImageNet we find that MSE is a substantal improvement over KL-divergence for shorter training (22, 45 epochs), but performs slightly worse for for longer training. However we observe that MSE has very little observable impact to BERT pretraining.

\subsection{Turn it off: Early-phase-only distillation is optimal for BERT but not ResNet-50}

\begin{figure}[ht]
    \centering
    \includegraphics[width=\linewidth]{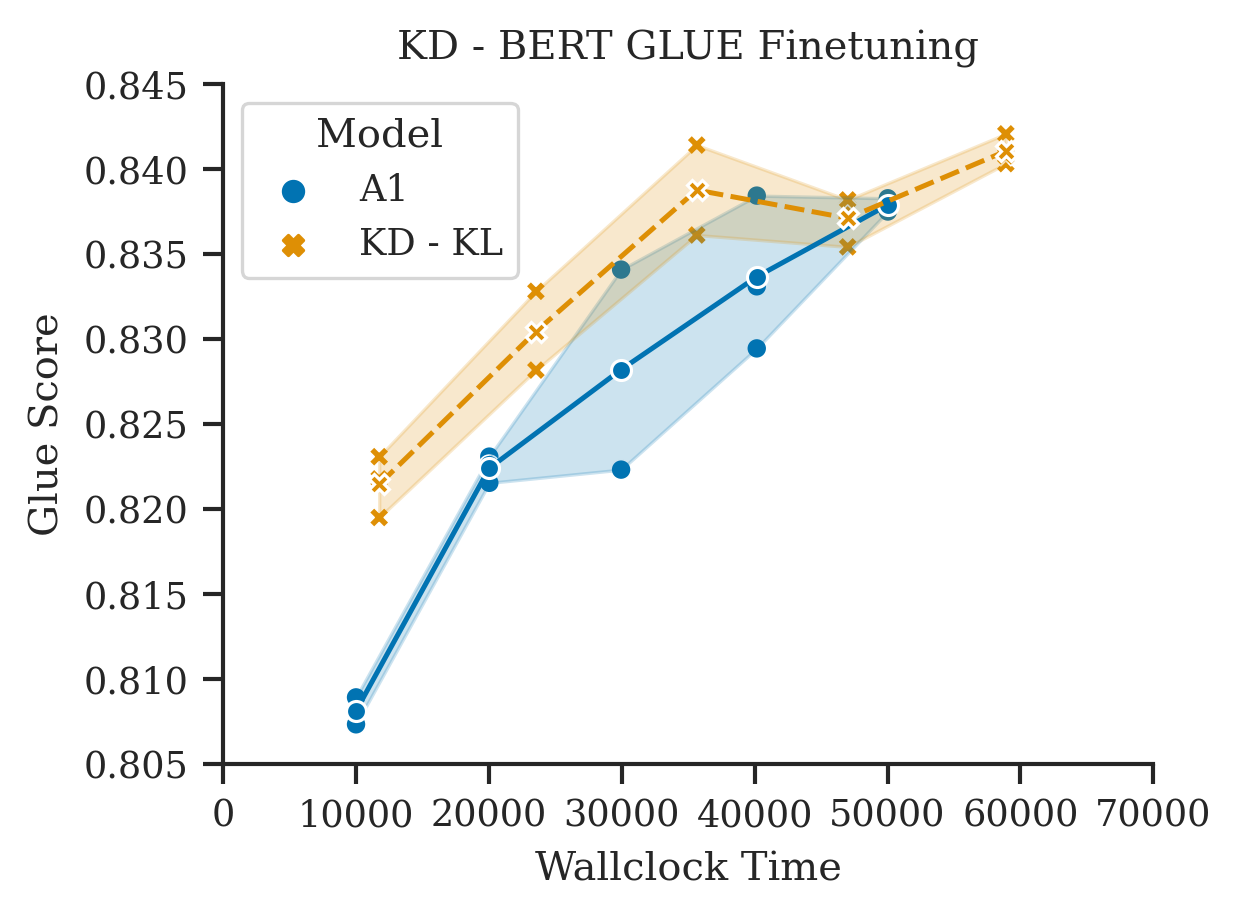}
    \caption{GLUE score of BERT pretrained on C4 dataset when applying the optimization of stopping distillation early. We observe that when trained for the full duration as the teacher model there is a slight dip in GLUE score. Note that distillation was only used during pretraining; GLUE finetuning was performed without distillation.}
    \label{fig:bert_glue_score}
\end{figure}

\begin{figure*}[ht]
    \centering
    \includegraphics[width=\linewidth]{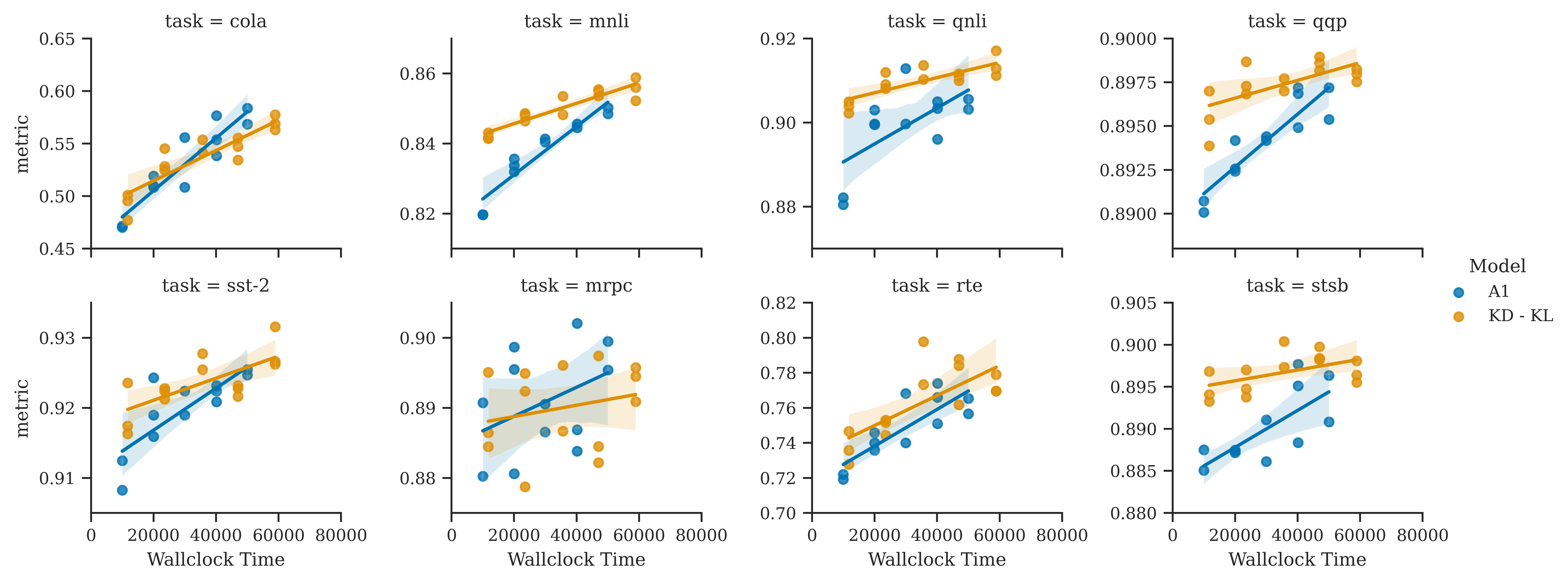}
    \caption{Wallclock comparisons of individual GLUE tasks for the highest quality baseline model (A1) and distilled model using A1 as a teacher (KD - KL) when pretraining for different durations. Note that the x-axis denotes the pretraining duration, not the duration of training on GLUE. Benefits from distillation are not equal across all tasks. Generally tasks with larger finetuning datasets saw the most benefit.}
    \label{fig:bert_glue_task}
\end{figure*}


Our previous experiments demonstrated that distillation can improve training speed of ResNet-50 on ImageNet by 1.96x, but they did not show that distillation improves training efficiency for BERT on C4). However, it's possible that our distillation configuration could be improved to yield further efficiency gains. For example, it's conventional to distill for the entirety of training. However, it's possible that the benefits of distillation are greater during a certain phase of training. If this is the case, then it may be more efficient to distill only for a subset of training given the computational cost of distillation.

Motivated by work highlighting the outsize impact of interventions applied during the early phase of neural network training \cite{gur-ari_gradient_2018,achille2018critical,sagun_empirical_early_2018,golatkar_time_early_2019,frankle_linear_2020,frankle_early_2020} and preliminary results implying the viability of scheduled distillation for reducing computational costs \cite{liu_knowledge_efficient_2022}, we conducted a series of experiments to determine whether scheduling distillation could yield further training efficiency improvements. In these experiments we allow of student models to train for a percentage of the total training duration (e.g 25\%, 50\%) then stop distilling.


In ResNet-50 trained on ImageNet, we found that stopping distillation early for Imagenet is never an efficiency improvement over distilling the whole time.

We also performed these experiments on BERT trained on C4. Interestingly, we found that stopping distillation early for BERT \emph{always} results in better or equal MLM Acc than leaving distillation on for the duration of training. Although the specific optimal for any training duration and teacher model tends to vary we found that the best percentage for training with distillation tended to fall with in the range of 15-40\% (see Appendix Figure \ref{fig:when-to-stop-distill}). 








After deciding on a suitable percentage of training for which to distill (30\%) we then trained again for our set of training duration. We find that we see reach the same accuracy as the baseline teacher model 1.20x (Figure \ref{fig:bert-pareto}) faster when applying distillation. The student model also reaches equal GLUE score as the baseline in 1.42x as fast (Figure \ref{fig:bert_glue_score}). 

The effects of distillation on downstream tasks is not spread evenly. In Figure \ref{fig:bert_glue_task} Applying distillation on the early phase of training allowed QNLI, and STSTB to match or exceed the baseline model in only 25\% of the training steps. MNLI, QQP, SST-2 were able to match or exceed the teacher model in 50\% of the training steps. Generally we observe that tasks with smaller and less stable datasets benifited less from distillation and tasks with larger datasets benefited more.

These results demonstrate that distillation schedules matter. Distillation for BERT yields optimal results when it is only performed for the first 20-50\% of training; training BERT with distillation for the entirety of training actually \emph{decreases} efficiency. In contrast, distilling for the entirety of training is optimal for ResNet-50 on ImageNet.





\subsection{Those who cannot do, teach: Sub-optimal models can be ideal teachers}
\label{sec:bad-for-teacher}

\begin{figure*}[ht]
     \centering
     \begin{subfigure}[t]{0.48\textwidth}
         \centering
         \includegraphics[width=\textwidth]{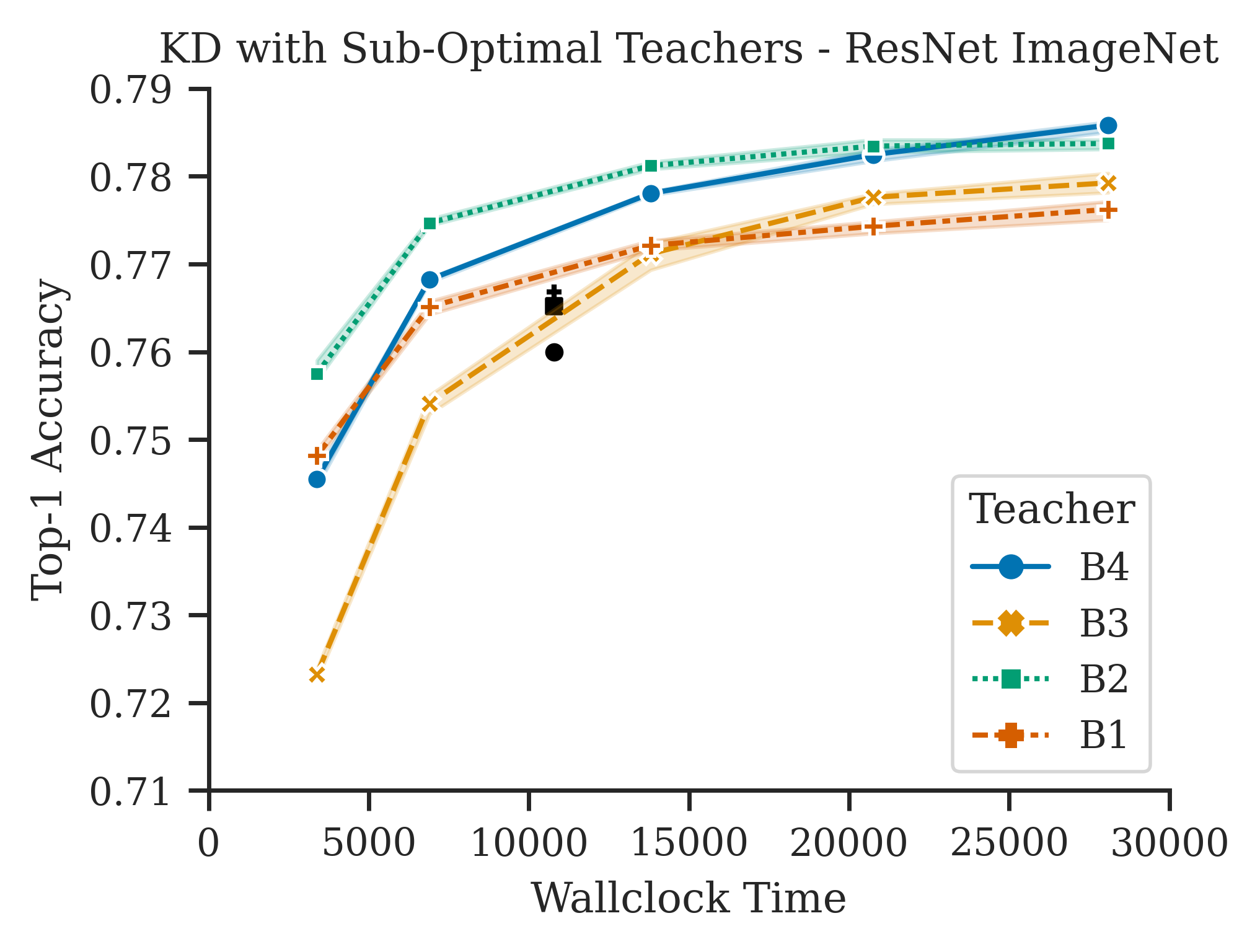}
        \caption{For ResNet models we see that not only is the best choie of teacher \emph{not} the teacher with the highest accuracy, but in fact which teacher is best depends on the training duration allotted.}
        \label{fig:resnet-sub-optimal}
     \end{subfigure}
     \hfill
     \begin{subfigure}[t]{0.48\textwidth}
         \centering
         \includegraphics[width=\textwidth]{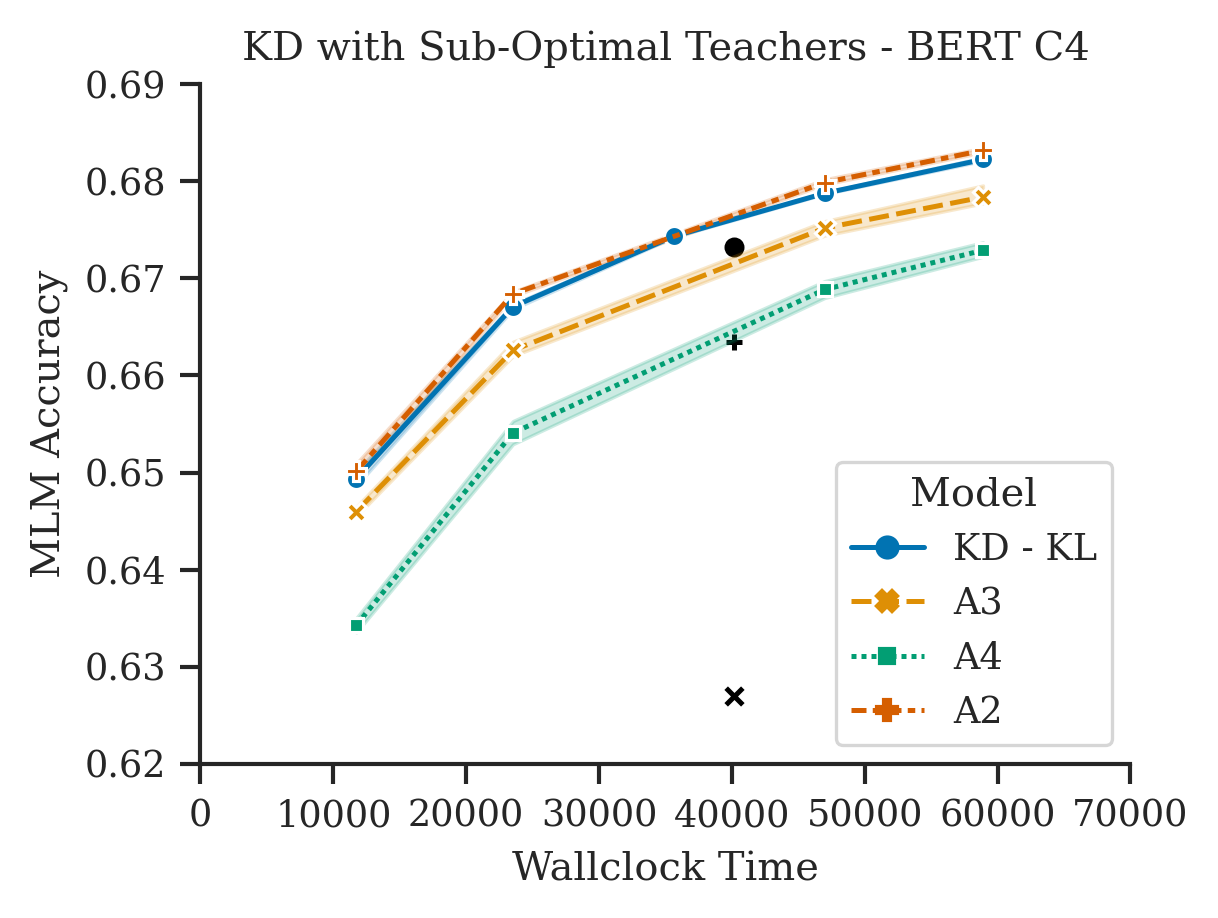}
         \caption{Wallclock MLM accuracy of BERT models trained with different teacher models. Teacher model accuracies and wallclock times are marked in black with matching marker. (A4 omitted for scaling)}
         \label{fig:bert-sub-optimal}
     \end{subfigure}
        \caption{Wallclock time vs accuracy plots for distilling ResNet and BERT using teachers with lower accuracy. For both domains we see the best teacher model is \emph{not} the teacher with the highest accuracy.}
        \label{fig:sub-optimal-joint}
\end{figure*}




We have demonstrated that that for both ResNet and BERT pre-training that using distillation can result in training efficiency gains. However, our previous experiments used the teacher that results in the best student model. However, models trained in a hyperparameter sweep will typically be of varying quality. While there are varying results regarding the correlation between the quality of a model and the quality of the same model as a teacher \cite{kaplun_bad_models_2022}, it's unclear whether sub-optimal models can improve training efficiency. Are the lower quality models simply an irredeemable waste of compute? If we select a model with poor accuracy can it still improve wall-clock efficiency, or do we run the risk of an outcome worse than if did not distill at all?

To investigate the impact of model quality on training efficiency, we designed a simple extension to the experiment in 
\ref{sec:simple}. For both ResNet and BERT we repeat the process of training with distillation for the same set of training durations, but this time with each possible teacher model from the hyperparameter sweep. If sub-optimal teacher models consistently have a negative impact on training efficiency, it's possible that training with distillation could take longer to reach the accuracy of a teacher model than training without distillation.

We present our results in Figure \ref{fig:sub-optimal-joint}. On ImageNet we find that three of our four possible teacher models yield efficiency improvements when training with distillation; i.e. three of the models trained with distillation reach the accuracy of the teacher models faster than the teacher models. Additionally none of the teacher model result in a step-wise improvement. We also find there appears to be little correlation between accuracy of the models and how they perform as teachers. Our best model from Section \ref{sec:simple} (B1) is the lowest performing teacher on 2 of 5 training durations (135, 180). Additionally, our third best model of the sweep (B4) becomes more competitive as the training duration increases, becoming the best choice teacher model at 180 epochs.

For masked language modeling pre-training on BERT we see simlar outcomes (Figure \ref{fig:bert-sub-optimal}). Two of the models trained with distillation reach the accuracy of the best non-distilled model (A1) faster than the best non-distilled model, and \emph{all} of the models trained with distillation reach the accuracy of the remaining three non-distilled models faster than the non-distilled models themselves. We also see again that the highest quality non-distilled model (A1) is not the best teacher (A2). Additionally, the student taught with lowest-quality non-distilled model (A4) ultimately reaches an MLM accuracy that is nearly 10 percentage point higher than its teacher at equal wallclock time. 

\begin{figure*}[ht]
    \centering
    \includegraphics[width=0.8\linewidth]{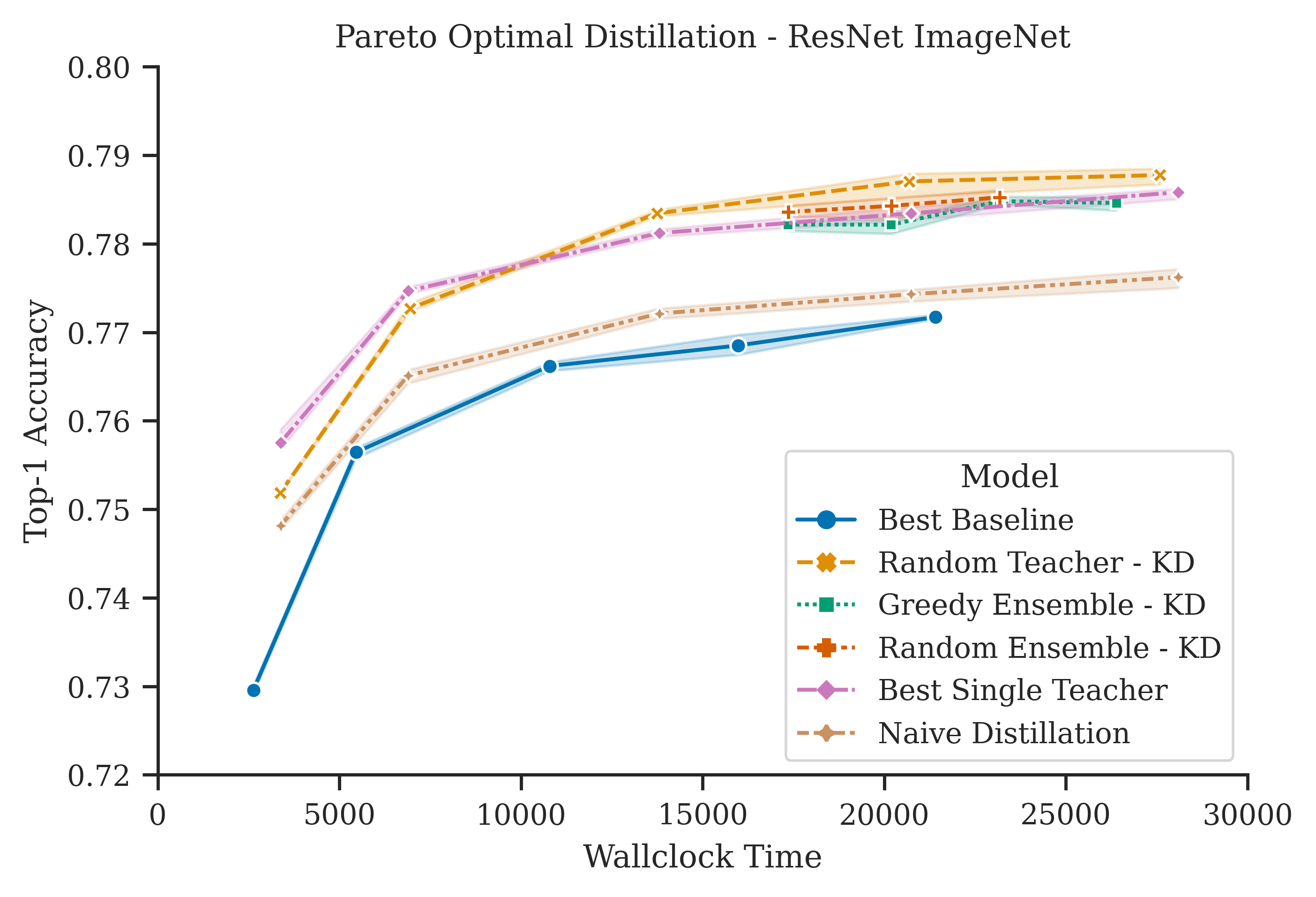}
    \caption{Pareto curve comparing single model distillation, multi model distillation, and randomly sampling teachers. (dark green baseline (no kd), blue single model distillation, orange 90 epochs 2, 3, 4,teachers chosen greadly (MSE), light green (90 epochs) 2, 3, 4, random selected teachers from 5 (MSE), pink randomly select 1 teacher from 5 ssr 0.25, 0.5, 1.0, 1.5, 2.0}
    \label{fig:pareto_random}
\end{figure*}

From the results we can make the following conclusions:  

\begin{enumerate}
    \item The risk of a training outcome worse than training without distillation is very low. Even when selecting the worst teacher, distillation is likely to be an efficiency improvement.
    \item The quality of a model does not consistently predict the quality of that model as a teacher for distillation. 
\end{enumerate}

Given that this is the case perhaps it is best \emph{not} to choose what the right teacher is in the first place.

\subsection{Optimal performance through random teacher sampling}

In the previous experiments we only used a single teacher. We showed that distillation usually improves training efficiency, even with low-quality teachers. However, using a single teacher still fails to leverage the resources spent on training the remaining models that \emph{are not} used as teachers. In the spirit of the Iterated Runs Problem, we will attempt to leverage the benefits of all our trained models.

We look for inspiration to model ensembling to gain additional model reuse. Unfortunately, training with multiple teachers increases the computational overhead proportionally to the number of teachers, making it unclear whether using multiple teachers is an efficiency improvement. Preliminary research on ResNet56 trained on CIFAR100 has demonstrated stepwise improvements when distilling from ensembles of teachers as well as randomly sampling subsets of the at each step \cite{liu_knowledge_efficient_2022}. In this section we investigate how leveraging multiple teachers can lead to further efficiency improvements. 

It's possible that \emph{consistency} is important and that the static and constant response provided by a greedily selected ensemble of teachers will be the most effective (as noted in \cite{beyer_patient_distillation_2021}). Alternatively, it's possible that sampling from a larger population of \emph{diverse} teachers is the most efficient configuration. We also want to examine whether additional teachers improve training efficiency. It's possible that the quality improvement from adding more teachers outweighs the additional computational burden.

\subsubsection*{Setup}

We study the efficiency of teacher ensembles for distillation in the following three paradigms:

\begin{itemize}
    \item By building ensembles in a greedy fashion.
    \item By building ensembles on the fly at each training step by randomly sampling a subset of teachers.
    \item By sampling only a single teacher model at random.
\end{itemize}

Our first method selects the models with best accuracy one at a time. Testing all possible combinations of teachers would be computationally prohibitive, but scaling up the size of our teacher ensembles in a greedy fashion by selecting first from the models with the highest accuracy is a reasonable and practical approach to building such an ensemble. 

Our second method ensembles on-the-fly by loading all models onto gpu memory, then choosing the desired number at random without replacement on each step and averaging their responses. This does slightly impact hardware performance as more memory must be used even for the models not selected on a given step. 

Our third method samples only a single teacher model at random but varying the training duration. Its possible that exposure to the same variety of models is a sufficient approximation of a larger ensemble without the additional computational overhead.

In these experiments we again use the same set of teacher models from the previous section. We also add an additional teacher model trained with mix-up to improve model diversity (see Appendix Figure \ref{fig:imagenet-sub-optimal-m1} for comparison).  We now average the ensemble responses such that our new loss term is:

$$\mathcal{L} = \lambda \mathcal{L}_{mse}(z_t, z_s) + \mathcal{L}_{ce}$$

Where $z_{s}$ is the student logit vector and  $z_{t} = \frac{1}{n}\sum_{1}^{n}z_n$ is the mean teacher logit vector for $n$ number of teachers.

\subsubsection*{Baselines}
We compare these models to a selection of results from our previous section:

\begin{itemize}
    \item A naive baseline distilled from the model with the best accuracy (Section \ref{sec:simple}.)
    \item The top performing model from each training duration presented in Section \ref{sec:bad-for-teacher}
    \item The top performing teacher model from each training duration \ref{tab:teacher-val-naive-kd} 
\end{itemize}

The naive approach distills the model with the best reported accuracy at 90 epochs of training on ImageNet. The baselines from Section \ref{sec:bad-for-teacher} serve as a stronger baseline which would be impractical and computationally prohibitive for most applications.

\subsubsection*{Conclusion}

For all our experiments we observe an efficiency improvement over our baseline model (Figure \ref{fig:pareto_random}. In shorter training regimes (22 and 45 epochs) we find that selecting the best model from a single teacher performs best (Best Single Teacher). However, this approach requires one to exhaustively search the best teacher not just for the task, but also that training duration (as we have shown in Section \ref{sec:bad-for-teacher}). We also find distilling from an greedy ensemble of teacher models is not a pareto improvement as compared to selecting the best single teacher, but randomly sampling the ensemble on the fly \emph{is}.

We find that random sampling a single teacher per step from an ensemble of teachers is able to reach the Top-1 Accuracy of our best baseline model trained for 90 epoch \textbf{1.85}x faster. For all range of values we find that sampling a teacher at random for distillation resulted in a speedup of \textbf{1.32}-\textbf{3.16}x for all training duration. This approach


We also find that the choice of loss function is not trivial. When using KL Divergence the inclusion of certain models hurt performance. Those models seemed to drag the accuracy down to be no better than that of the worst single model distillation performance. MSE on the other hand, while never as good at distilling with only a single teacher was much more robust when adding models with poor performance.

\section{Discussion}

We conducted a series of experiments to investigate the utility of distillation for improving training efficiency using ResNet-50 trained on ImageNet and BERT trained on C4 and evaluated on GLUE. We found that distillation improves training efficiency: it can speed up training by up to 1.96x in ResNet-50 trained on ImageNet and up to 1.42x on BERT when evaluated on GLUE. We also found that distillation schedules matter. Distillation for BERT yields optimal results when it is only performed for the first 20-50\% of training, but that distilling for the entirety of training is optimal for ResNet-50 on ImageNet. Furthermore, we found that model quality does not consistently predict teacher quality. Training with distillation is almost always more efficient than training without distillation, even when using the poorest-quality model as a teacher, in both ResNet-50 and BERT. We were also able to reduce the runtime cost of teacher ensembles from $O(N)$ to $O(1)$ while still retaining their benefits to distillation by randomly sampling one teacher model from a pool of teachers on each iteration. Finally, we observed differences between mean squared error (MSE) and KL-Divergence (KL) distillation loss in ResNet-50 trained on ImageNet. MSE is more robust—it more consistently yields higher quality student models across a wide range hyperparameter values—but KL-Divergence distillation loss yields the \emph{best} student models.

One caveat to our work is that distillation requires loading a teacher model into GPU memory. Depending on the size of the teacher and student models and the amount of GPU memory, distillation can exceed the GPU memory capacity. In such a scenario, memory-saving techniques such as gradient accumulation may be necessary, which can impose additional computational overhead. Accordingly, distillation may no longer improve efficiency such a scenario. This emphasizes the value of future work examining strategies for reducing the memory overhead of distillation, for example by caching teacher outputs or using smaller teacher models.

Future work could also explore to what extent BERT can take advantage of ensembles of previous teacher models. GPU memory constraints made direct extensions with our ResNet experiments difficult.

Another shortcoming of this work is that we do not provide a precise nor analytical basis for our recommendation about when to stop using distillation when pretraining BERT. Our recommendation of stopping 20-50\% through training is derived entirely from empirical observation.



Our findings demonstrate that distillation consistently improves training efficiency in both image classification and language modeling across a range of training durations and teacher model qualities. We also show that the benefits of distillation on training speed and model quality are fungible, meaning that our proposed optimizations to distillation protocols—randomly sampling from ensembles of teacher models, and distilling for the beginning ~30\% of training (in BERT)—can be flexibly leveraged to reduce training time or increase model quality, depending on the needs of the practitioner. Taken together, this work emphasizes the value of distillation for improving the efficiency of training deep neural networks.

\bibliography{ref}
\bibliographystyle{mlsys2023}

\newpage 
\appendix
\section{Appendix}

\begin{table*}
\caption{Results of hyper parameter sweep of teacher models on both ResNet-50 ImageNet and BERT on C4 and KD. The teacher model is the highest-quality model at standard training length (B1 for ResNet-50, A1 for BERT). KL: KL-Divergence distillation loss; MSE: Mean-squared error distillation loss.}
\label{tab:teacher-val-naive-kd}
\centering
\adjustbox{max width=\linewidth}{%
\begin{tabular}{llrrrrrr}
\toprule
               & Model &            B1 &            B2 &            B3 &            B4 &       KD - KL &      KD - MSE \\
{ResNet-50 - ImageNet} & Epochs &               &               &               &               &               &               \\
\midrule

\multirow{5}{*}{Wallclock Time} & 22  &   2633.596938 &   2633.596938 &   2633.596938 &   2633.596938 &   3377.949392 &   3384.856735 \\
                & 45  &   5442.526072 &   5442.526072 &   5442.526072 &   5442.526072 &   6892.535721 &   6904.456440 \\
                & 90  &  10792.137267 &           NaN &  10792.137267 &  10792.137267 &  13802.963346 &  13723.052594 \\
                & 135 &  15966.039060 &  15966.039060 &  15966.039060 &  15966.039060 &  20743.611829 &  20550.213244 \\
                & 180 &  21418.041385 &  21418.041385 &  21418.041385 &  21418.041385 &  28181.851872 &  27523.290798 \\
\bottomrule
\end{tabular}}
\end{table*}

\begin{table*}[]
    \centering
        \caption{Table for figure \ref{fig:bert-pareto}}
\begin{tabular}{llrr}
\toprule
         &       &  Wallclock Time &  MLM Accuracy \\
Teacher & Training Steps &                 &               \\
\midrule
\multirow{5}{*}{A1} & 17500 &    11736.988181 &      0.649377 \\
         & 35000 &    23534.059697 &      0.667094 \\
         & 52500 &    35640.971302 &      0.674340 \\
         & 70000 &    46999.459847 &      0.678768 \\
         & 87500 &    58905.076343 &      0.682252 \\
\cline{1-4}
\multirow{5}{*}{baseline} & 17500 &     9994.233387 &      0.631140 \\
         & 35000 &    20025.788366 &      0.654882 \\
         & 52500 &    29945.809139 &      0.665876 \\
         & 70000 &    40156.393939 &      0.673092 \\
         & 87500 &    50021.559775 &      0.677736 \\
\bottomrule
\end{tabular}

    \label{tab:bert-stop-table}
\end{table*}

\begin{table*}[]
    \centering

\caption{Table of mean wallclock time per seed, training steps and glue score for figure \ref{fig:bert_glue_score}}
\begin{tabular}{llrr}
\toprule
        &         &  Wallclock Time &  Glue Score \\
Model & steps &                 &             \\
\midrule
\multirow{5}{*}{A1} & 17500.0 &     9994.233387 &    0.808114 \\
        & 35000.0 &    20025.788366 &    0.822399 \\
        & 52500.0 &    29945.809139 &    0.828202 \\
        & 70000.0 &    40156.393939 &    0.833666 \\
        & 87500.0 &    50021.559775 &    0.837898 \\
\cline{1-4}
\multirow{5}{*}{KD - KL} & 17500.0 &    11736.988181 &    0.821447 \\
        & 35000.0 &    23534.059697 &    0.830443 \\
        & 52500.0 &    35640.971302 &    0.838772 \\
        & 70000.0 &    46999.459847 &    0.837115 \\
        & 87500.0 &    58905.076343 &    0.841064 \\
\bottomrule
\end{tabular}

\quad

    \caption{Table with glue results per task for figure \ref{fig:bert_glue_task}. Training Steps for teacher model and shortest training duration result to match or exceed teacher marked in bold}
    \label{tab:glue-table}

    \centering

\begin{tabular}{llrrrrrrrr}
\toprule
        & {} & \multicolumn{8}{l}{Task} \\
        &  &      cola &      mnli &      mrpc &      qnli &       qqp &       rte &     sst-2 &      stsb \\
Model & Steps &           &           &           &           &           &           &           &           \\
\midrule
\multirow{5}{*}{A1} & 17500 &  0.470822 &  0.819690 &  0.885496 &  0.881292 &  0.890398 &  0.720578 &  0.910359 &  0.886278 \\
        & 35000 &  0.512419 &  0.833799 &  0.891598 &  0.900726 &  0.893056 &  0.740554 &  0.919725 &  0.887317 \\
        & 52500 &  0.532160 &  0.840914 &  0.888541 &  0.906279 &  0.894295 &  0.754152 &  0.920680 &  0.888591 \\
        & \textbf{70000} &  0.556154 &  0.844962 &  0.890907 &  0.901458 &  0.896321 &  0.763658 &  0.922146 &  0.893724 \\
        & 87500 &  0.576097 &  0.849339 &  0.897433 &  0.904357 &  0.896283 &  0.761011 &  0.925076 &  0.893591 \\
\cline{1-10}
\multirow{5}{*}{KD - KL} & 17500 &  0.491222 &  0.842059 &  0.888649 &  \textbf{0.903716} &  0.895416 &  0.736703 &  0.919088 &  \textbf{0.894724} \\
        & 35000 &  0.532878 &  \textbf{0.847676} &  0.888682 &  0.909696 &  \textbf{0.897594} &  0.749699 &  \textbf{0.922146} &  0.895175 \\
        & 52500 &  0.547583 &  0.850865 &  0.\textbf{891368} &  0.911953 &  0.897360 &  \textbf{0.785560} &  0.926606 &  0.898884 \\
        & 70000 &  0.545607 &  0.854683 &  0.888026 &  0.910855 &  0.898576 &  0.777858 &  0.922528 &  0.898785 \\
        & 87500 &  \textbf{0.569853} &  0.855701 &  0.893711 &  0.913723 &  0.897911 &  0.772804 &  0.928135 &  0.896673 \\
\bottomrule
\end{tabular}

    \label{tab:my_label}
\end{table*}

\begin{figure*}[h]
    \centering
    \includegraphics[width=.7\linewidth]{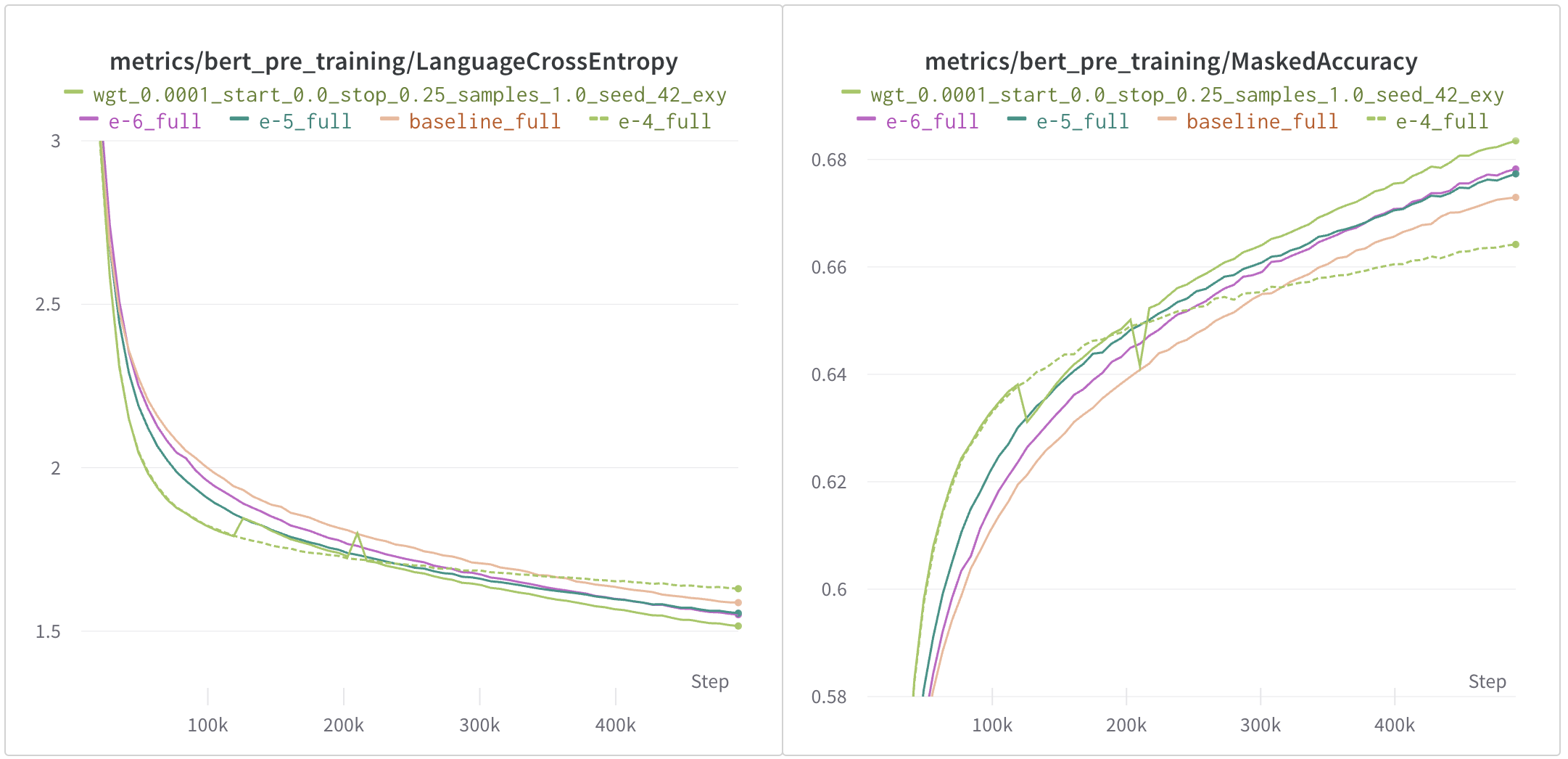}
    \caption{Comparison of learning rates and stopping points for distilling BERT. "Turn it off"}
    \label{fig:bert-stop-distill}
\end{figure*}

\begin{figure*}[h]
    \centering
    \includegraphics[width=.7\linewidth]{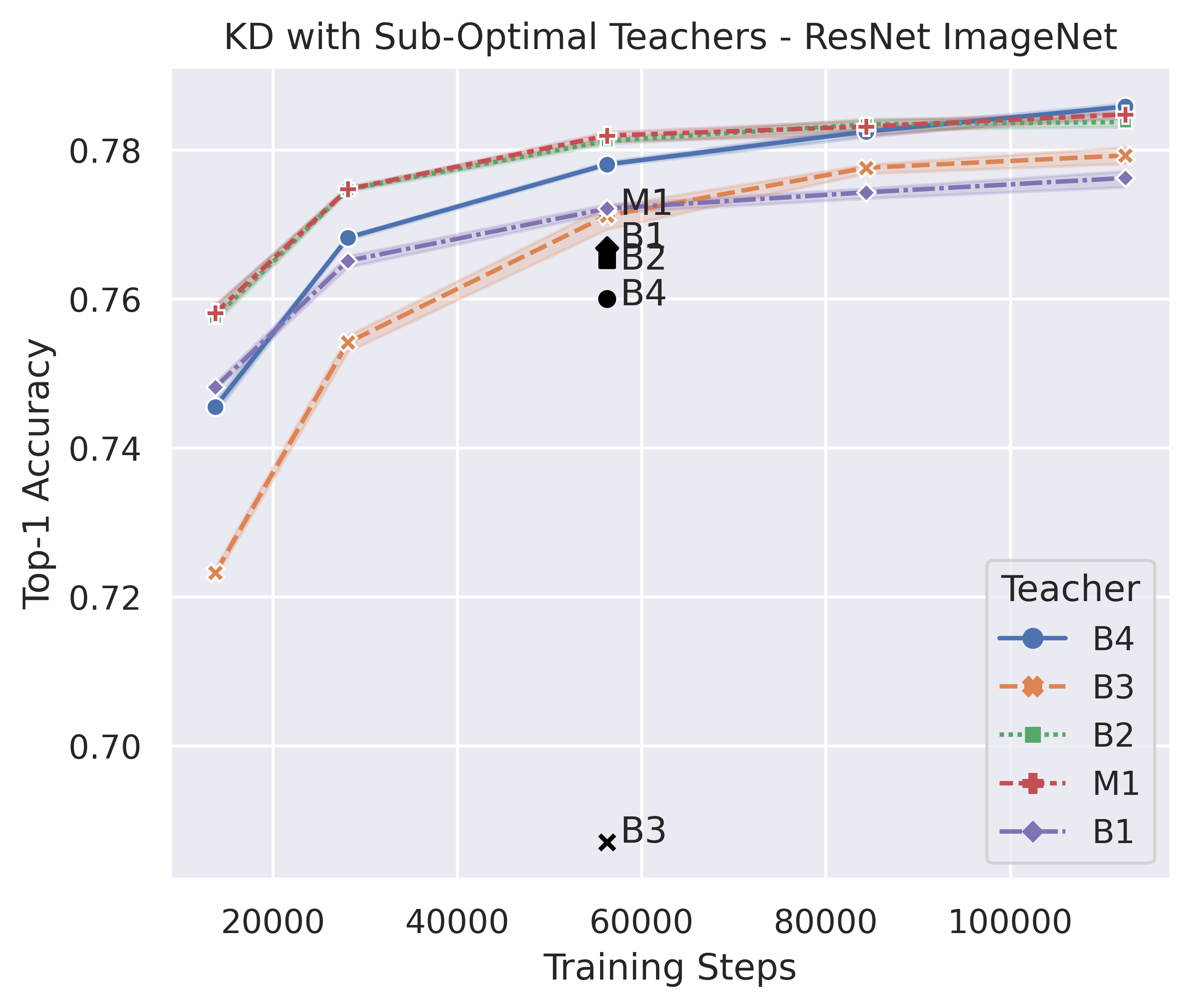}
    \caption{Comparison distillation with suboptimal teachers including teacher trained with mixup (M1). Accuracies of teachers marked in black.}
    \label{fig:imagenet-sub-optimal-m1}
\end{figure*}

\begin{figure*}[ht]
     \centering
     \begin{subfigure}[b]{0.48\textwidth}
         \centering
         \includegraphics[width=\textwidth]{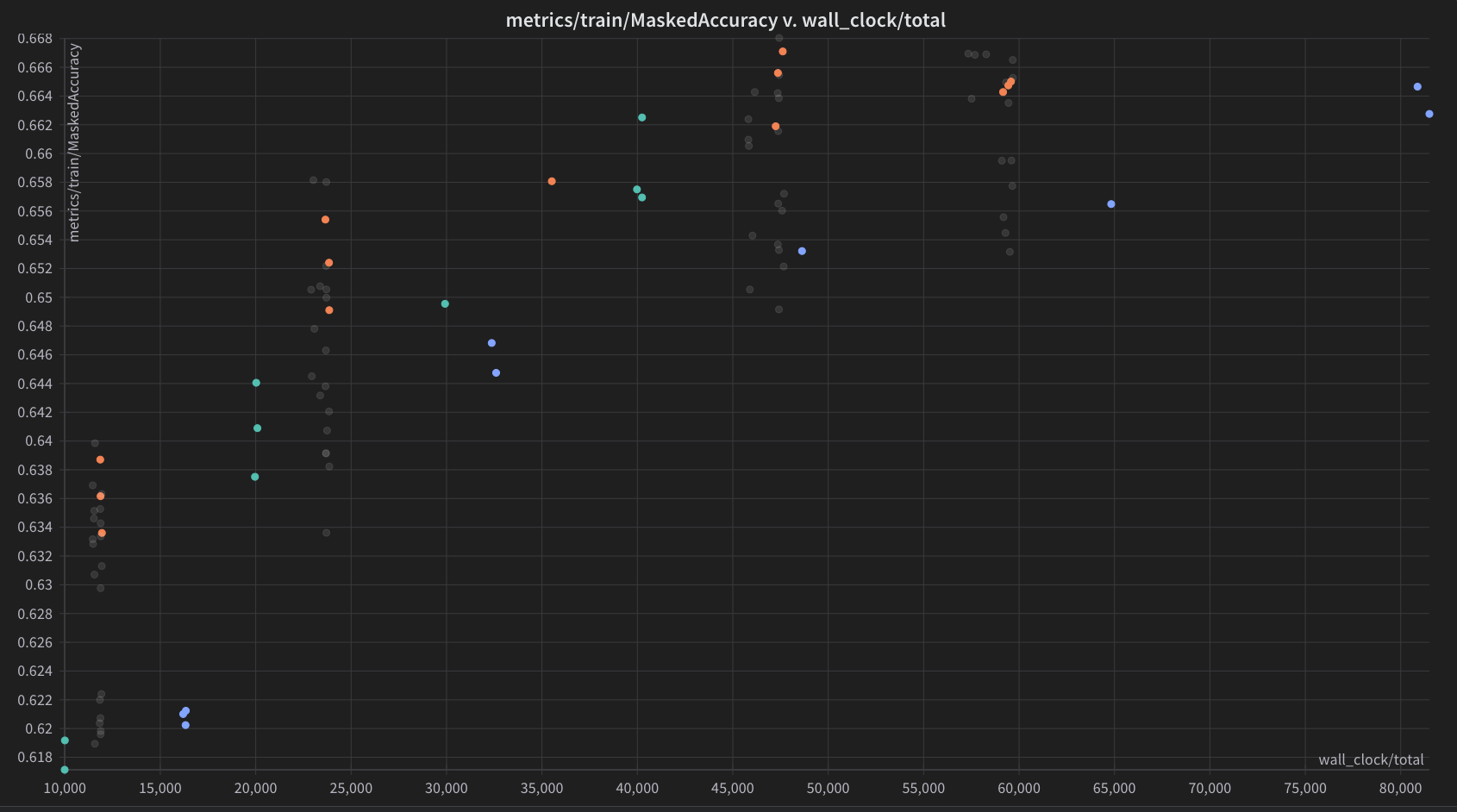}
         \caption{wallclock MLM accuracy of BERT models.}
         \label{fig:wallclock-turn-it-off}
     \end{subfigure}
     \hfill
     \begin{subfigure}[b]{0.48\textwidth}
         \centering
         \includegraphics[width=\textwidth]{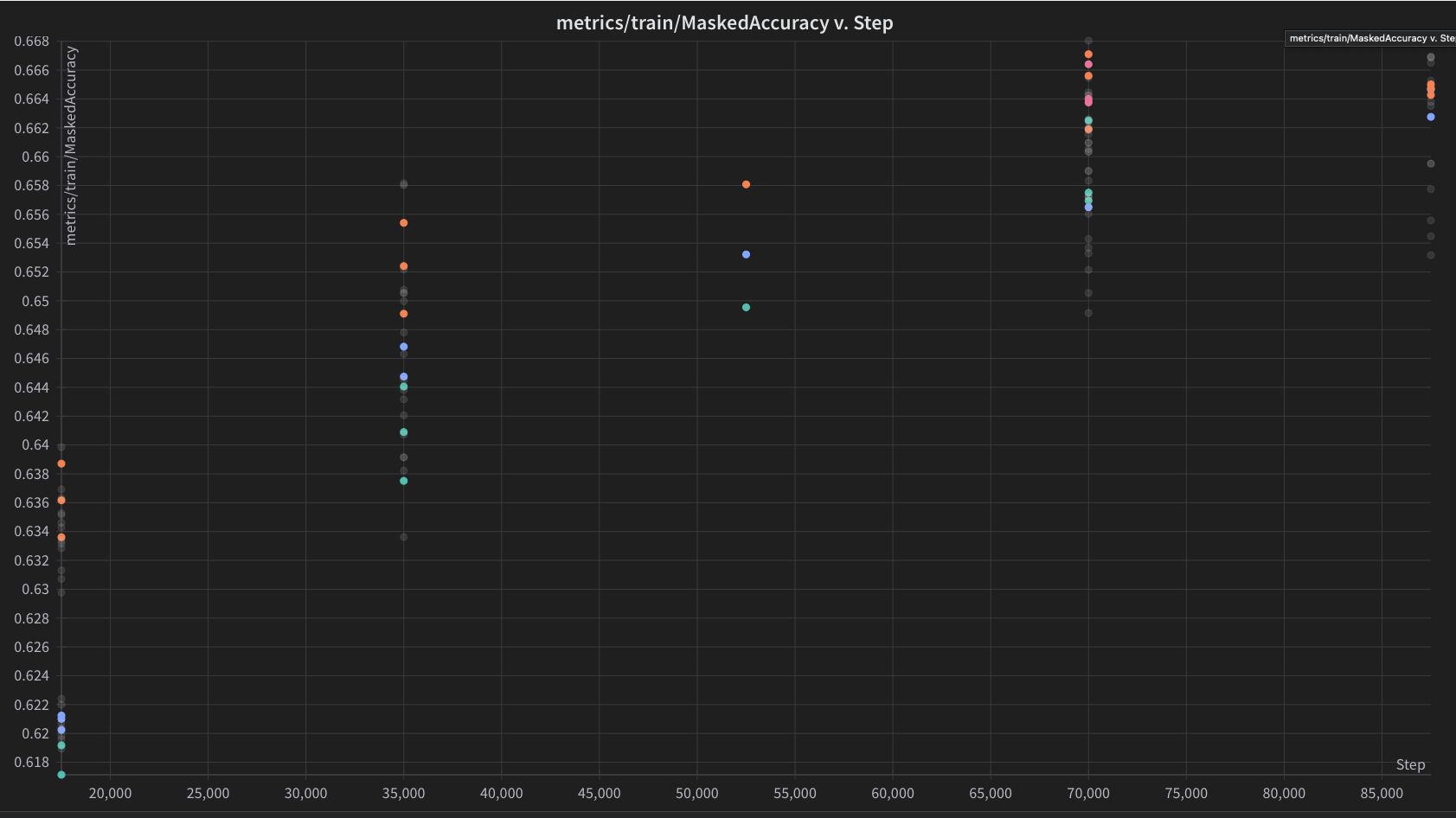}
         \caption{Stepwise MLM accuracy of BERT models.}
         \label{fig:stepwise-turn-it-off}
     \end{subfigure}
        \caption{ Pareto curve comparing Baseline BERT, distilling for all of training, and stopping distilling. (light green baseline (no kd), blue  distillation for all of training, orange stopping distillation at 30\% of training samples.}
        \label{fig:lofi-bert-charts}
\end{figure*}

\begin{figure}[h]
    \centering
    \includegraphics[width=\linewidth]{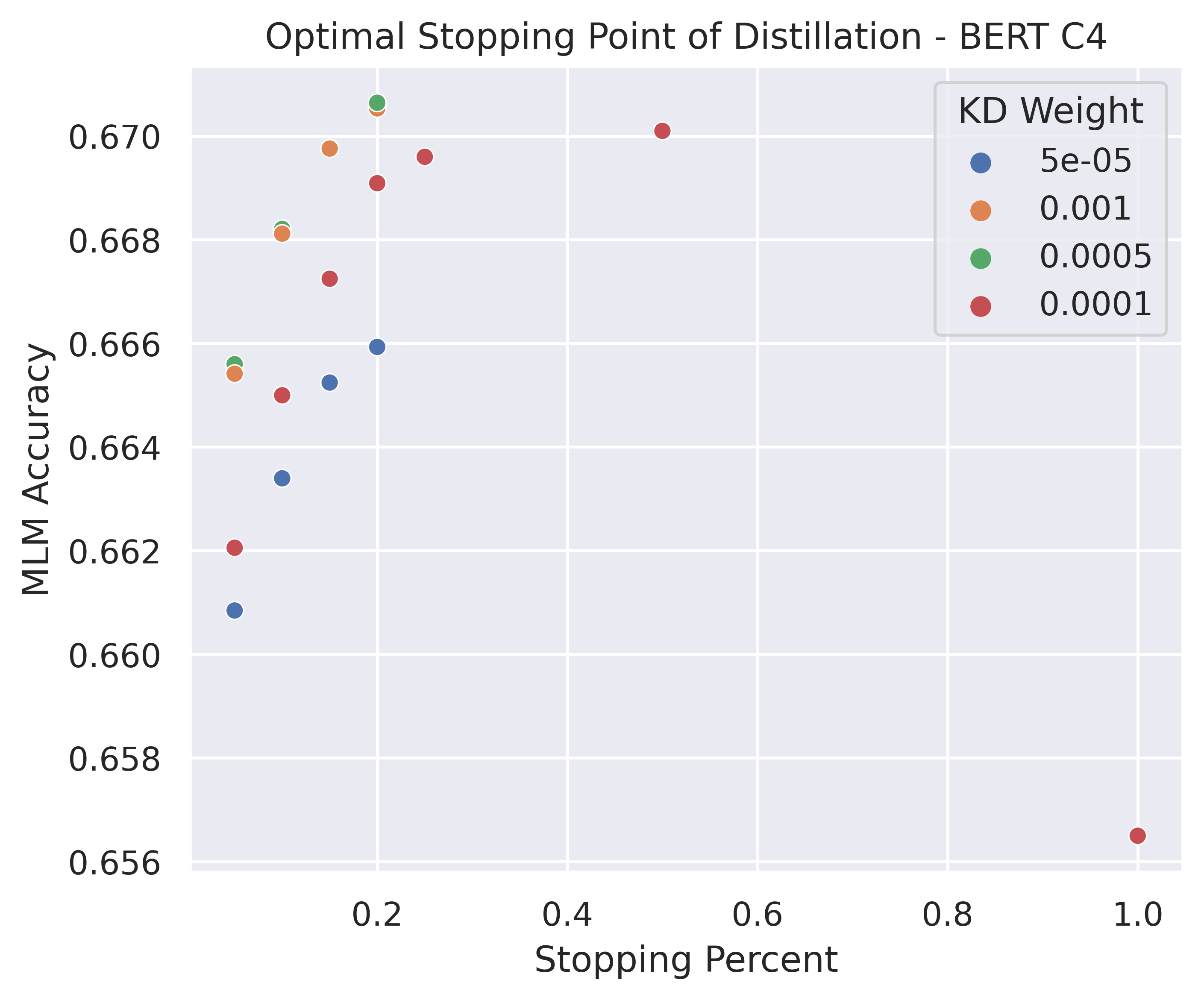}
    \caption{When to "Turn it off"}
    \label{fig:when-to-stop-distill}
\end{figure}


\end{document}